\documentclass[conference,a4paper]{IEEEtran}
\IEEEoverridecommandlockouts

\usepackage[hidelinks]{hyperref}
\usepackage[cmex10]{amsmath}%American Math Society(AMS) math formatting
\usepackage{amssymb,amsfonts}%AMS extra symbols and fonts
\usepackage{dblfloatfix}%fix double column figure ordering and placement

\usepackage[ruled,vlined]{algorithm2e}
\usepackage{graphicx}
\graphicspath{{Figures/PDF/}{Figures/PNG/}}

\usepackage{booktabs}
\usepackage{siunitx}
\usepackage[numbers,compress]{natbib}
\usepackage{texnames}
\usepackage{bm,bbm}
\usepackage{orcidlink}

\begin{document}

\title{Satellite-to-Street: Synthesizing Post-Disaster Views\\
from Satellite Imagery via Generative Vision Models}

\author{
\IEEEauthorblockN{Yifan Yang\orcidlink{0009-0009-9496-6925}}
\IEEEauthorblockA{
\textit{Department of Geography} \\
\textit{Texas A\&M University} \\
College Station, TX, USA \\
yyang295@tamu.edu
}
\and
\IEEEauthorblockN{Lei Zou\orcidlink{0000-0001-6206-3558}}
\IEEEauthorblockA{
\textit{Department of Geography} \\
\textit{Texas A\&M University} \\
College Station, TX, USA \\
lzou@tamu.edu
}
\and
\IEEEauthorblockN{Wendy Jepson\orcidlink{0000-0002-7693-1376}}
\IEEEauthorblockA{
\textit{Department of Geography} \\
\textit{Texas A\&M University} \\
College Station, TX, USA \\
wjepson@tamu.edu
}
\thanks{Corresponding author: Lei Zou.}
}

\maketitle
\begin{abstract}
In the immediate aftermath of natural disasters, rapid situational awareness is critical. Traditionally, satellite observations are widely used to estimate damage extent. However, they lack the ground-level perspective essential for characterizing specific structural failures and impacts. Meanwhile, ground-level data (e.g., street-view imagery) remains largely inaccessible during time-sensitive events. This study investigates Satellite-to-Street View Synthesis to bridge this data gap. We introduce two generative strategies to synthesize post-disaster street views from satellite imagery: a Vision-Language Model (VLM)-guided approach and a damage-sensitive Mixture-of-Experts (MoE) method. We benchmark these against general-purpose baselines (Pix2Pix, ControlNet) using a proposed Structure-Aware Evaluation Framework. This multi-tier protocol integrates (1) pixel-level quality assessment, (2) ResNet-based semantic consistency verification, and (3) a novel VLM-as-a-Judge for perceptual alignment. Experiments on 300 disaster scenarios reveal a critical realism--fidelity trade-off: while diffusion-based approaches (e.g., ControlNet) achieve high perceptual realism, they often hallucinate structural details. Quantitative results show that standard ControlNet achieves the highest semantic accuracy ($F_1=0.71$), whereas VLM-enhanced and MoE models excel in textural plausibility but struggle with semantic clarity. This work establishes a baseline for trustworthy cross-view synthesis, emphasizing that visually realistic generations may still fail to preserve critical structural information required for reliable disaster assessment.\end{abstract}


\section{Introduction}
\begin{figure*}[t]
  \centering
  \includegraphics[width=\textwidth]{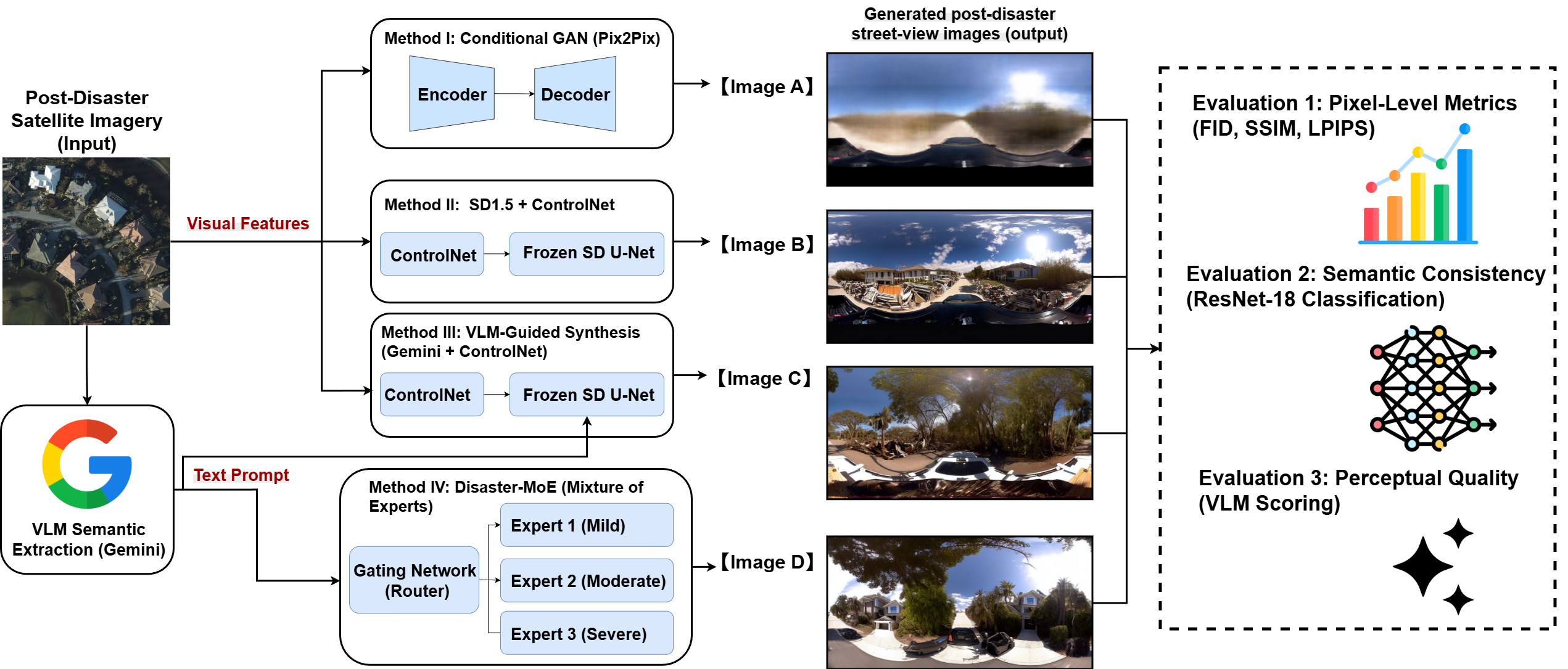}
  \caption{Overview of the proposed structure-aware evaluation framework and generative benchmarks. We systematically evaluate four paradigms: (A) Pix2Pix (baseline), (B) Stable Diffusion with ControlNet (baseline), (C) VLM-guided synthesis (Ours), and (D) Disaster-MoE framework (Ours). The generated street-view images are assessed using a multi-tier protocol: pixel-level metrics, ResNet-based semantic consistency, and VLM-driven perceptual quality.}
  \label{fig:method}
\end{figure*}

Disaster damage assessment relies heavily on imagery to capture the impacts of extreme events. While satellite imagery enables rapid and large-scale observation, its overhead perspective limits the visibility of critical side-view details, such as collapsed facades and debris. In contrast, street-view imagery provides human-scale context essential for structural assessment but is often scarce in post-disaster settings due to physical obstacles, including road blockages, debris, flooding, and restricted site accessibility~\cite{yang2026damagearbiter}. Although recent studies have demonstrated the utility of street-view data for hyperlocal damage classification~\cite{YANG2025102335, yang2025perceiving} and cross-view geolocalization~\cite{li2025cross}, systematic research on generating post-disaster street views directly from satellite imagery remains limited.

While Cross-View Image Synthesis (CVIS) has matured in urban computer vision~\cite{biljecki2021street}, extending it to disaster scenarios introduces unique challenges. First, traditional Generative Adversarial Networks (GANs), such as Pix2Pix~\cite{goodfellow2014generative}, often suffer from mode collapse in complex disaster scenes, resulting in blurred textures unsuitable for identifying wreckage details. Second, although recent diffusion-based models~\cite{ho2020denoising} demonstrate high fidelity, they are prone to structural hallucination in disaster contexts, referring to unintentionally "repairing'' damaged buildings rather than reproducing the actual destruction in disaster scenarios. Furthermore, the inherent imbalance between damaged and undamaged samples complicates the maintenance of semantic consistency.

To address these challenges, this study systematically compares general-purpose baselines (Pix2Pix, ControlNet) with two newly proposed disaster-adapted strategies: a VLM-guided approach that incorporates disaster-aware semantic prompts, and a Mixture-of-Experts (MoE) framework designed to account for damage-specific visual patterns. We introduce a Structure-Aware Evaluation Framework that integrates (1) traditional Image Quality Assessment (IQA) metrics, (2) ResNet-based semantic consistency verification, and (3) VLM-driven perceptual alignment. Using this protocol, we analyze the critical trade-off between structural fidelity and semantic consistency, establishing an empirical baseline for trustworthy cross-view synthesis in disaster response.

\section{Methodology}

\subsection{Problem Definition and Dataset}
The objective is to learn a mapping $G: I_{\text{sat}} \to I_{\text{street}}$ that synthesizes a ground-level view $I_{\text{street}}$ from a post-disaster satellite image $I_{\text{sat}}$. We utilize the 2022 Hurricane Ian dataset adapted from Li et al.~\cite{li2025cross}, comprising 4,121 paired satellite/street-view images. To ensure rigorous evaluation across damage severities, we constructed a balanced test set of 300 pairs, equally stratified into \textit{mild}, \textit{moderate}, and \textit{severe} damage levels. The remaining 3,821 pairs serve as the training set.

\subsection{Generative Frameworks}
As illustrated in Fig.~\ref{fig:method}, we evaluate four generative paradigms for synthesizing geometrically consistent and semantically aligned street-view images $\hat{I}_{\text{street}}$ from post-disaster satellite inputs $I_{\text{sat}}$.

\subsubsection{Method A: Pix2Pix (Conditional GAN)}
As a baseline for direct image-to-image translation, Pix2Pix learns a mapping $G: I_{\text{sat}} \to \hat{I}_{\text{street}}$ via adversarial training. The generator is optimized to minimize the combined adversarial and reconstruction objective:
\begin{equation}
\mathcal{L}_{\text{Pix2Pix}} = \mathcal{L}_{\text{GAN}}(G,D) + \lambda \| I_{\text{street}} - \hat{I}_{\text{street}} \|_1.
\end{equation}

\subsubsection{Method B: ControlNet-Guided Diffusion}
To improve visual fidelity, we adopt a Latent Diffusion Model (LDM) conditioned on satellite imagery. The forward process diffuses the latent street-view representation $\mathbf{z}_0$ into noise:
\begin{equation}
\mathbf{z}_t = \sqrt{\alpha_t}\mathbf{z}_0 + \sqrt{1-\alpha_t}\boldsymbol{\epsilon}, \quad \boldsymbol{\epsilon} \sim \mathcal{N}(0, \mathbf{I}).
\end{equation}
In the reverse process, ControlNet injects multi-scale spatial constraints $\mathcal{C}(I_{\text{sat}})$ into the frozen U-Net, guiding the noise prediction:
\begin{equation}
\boldsymbol{\epsilon}_\theta = \epsilon_\theta(\mathbf{z}_t, t \mid \mathcal{C}(I_{\text{sat}})).
\end{equation}
This ensures strong geometric alignment between the satellite layout and the generated view.

\subsubsection{Method C: VLM-Guided Synthesis}
To capture explicit damage semantics, we introduce linguistic guidance using a Vision-Language Model (VLM, specifically Gemini-2.5-Flash). The VLM extracts a textual damage description $\mathbf{p}$ from $I_{\text{sat}}$:
\begin{equation}
\mathbf{p} = \Phi_{\text{VLM}}(I_{\text{sat}}).
\end{equation}
The generation is then jointly conditioned on structural features and semantic prompts:
\begin{equation}
\boldsymbol{\epsilon}_\theta = \epsilon_\theta(\mathbf{z}_t, t \mid \mathcal{C}(I_{\text{sat}}), \mathbf{p}).
\end{equation}
This formulation enhances the synthesis of disaster-specific attributes (e.g., debris, collapsed roofs) often missed by visual features alone.

\subsubsection{Method D: Disaster-MoE}
To address heterogeneous damage patterns, we propose a Mixture-of-Experts (MoE) framework. We train $K$ specialized ControlNet experts $\{E_k\}_{k=1}^K$ for distinct severity levels (mild, moderate, severe). An adaptive routing network $R$ predicts gating weights based on satellite features:
\begin{equation}
\mathbf{w} = R(I_{\text{sat}}), \quad \sum_{k=1}^K w_k = 1.
\end{equation}
The final denoising step dynamically aggregates expert predictions:
\begin{equation}
\boldsymbol{\epsilon}_\theta = \sum_{k=1}^K w_k \, \epsilon_\theta^{(k)}(\mathbf{z}_t, t \mid \mathcal{C}_k(I_{\text{sat}})).
\end{equation}
By explicitly routing samples to severity-specific experts, Disaster-MoE minimizes the confusion between intact and damaged structures.

\subsection{Evaluation Protocol}

To comprehensively assess the proposed framework, we implement a structure-aware evaluation protocol across three dimensions: pixel-level fidelity, semantic consistency, and perceptual alignment.

\subsubsection{Tier 1: Pixel-Level Metrics}
We employ standard metrics to quantify visual quality: Structural Similarity Index Measure (SSIM) and Peak Signal-to-Noise Ratio (PSNR) measure low-level structural and luminance fidelity, while Learned Perceptual Image Patch Similarity (LPIPS) and Fréchet Inception Distance (FID) evaluate deep feature alignment and distributional distance between synthesized and real images.

\subsubsection{Tier 2: Semantic Consistency (CAS)}
To verify if generated images preserve damage semantics, we adopt the Classification Accuracy Score (CAS). A ResNet-18 classifier, initialized with ImageNet pre-trained weights, is fine-tuned on real post-disaster street-view images using the Adam optimizer (learning rate $1\times10^{-4}$, batch size 32) for 10 epochs. After training, the classifier parameters are frozen, and the model is used to predict damage severity levels on synthesized street-view images for semantic consistency evaluation. We report F1 scores and confusion matrices to quantify how well the generative models maintain severity-specific features (e.g., distinguishing \textit{severe} debris from \textit{mild} clutter).

\subsubsection{Tier 3: VLM-as-a-Judge}
Complementing traditional metrics, we utilize a Vision-Language Model (Gemini-2.5-Flash) to approximate human perceptual judgment. The model compares generated images against ground truth on a 5-point Likert scale across three criteria: (1) \textit{Structural Consistency} (layout alignment), (2) \textit{Damage Accuracy} (correctness of severity representation), and (3) \textit{Perceptual Realism} (visual plausibility). This tier captures high-level nuances often missed by pixel-wise metrics.

\section{Results and Discussion}
\begin{figure*}[t]
  \centering
  \includegraphics[width=\textwidth]{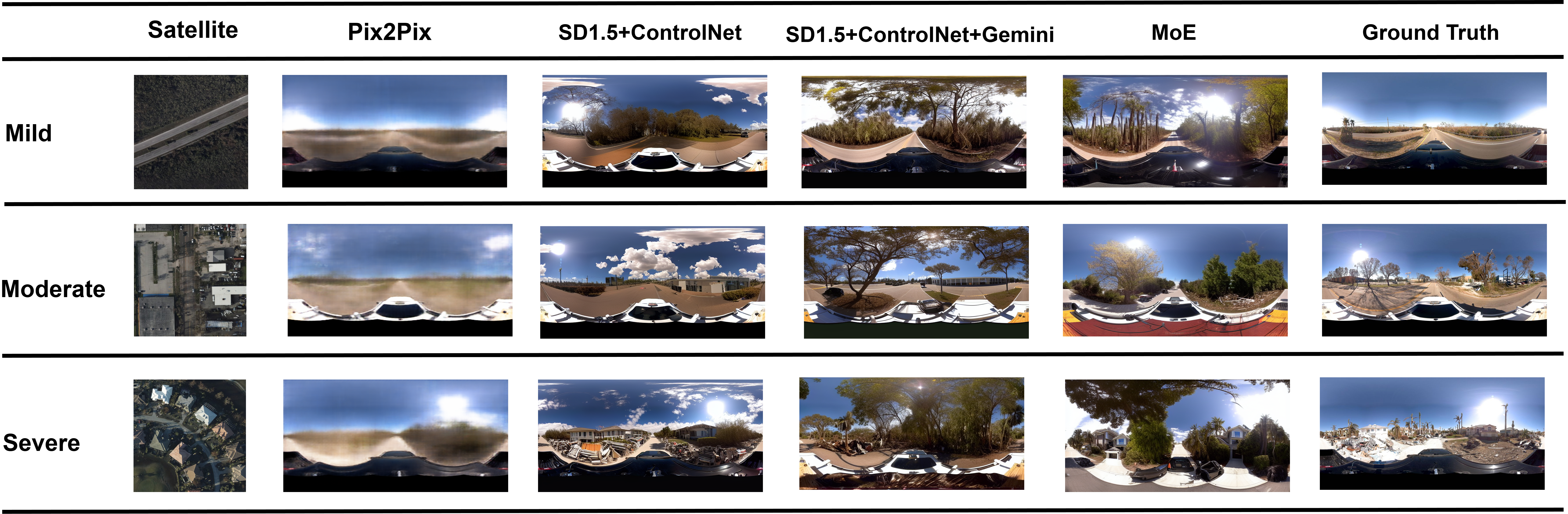}
  \caption{Qualitative comparison of satellite-to-street-view synthesis results across different disaster severity levels. Columns correspond to Pix2Pix, Stable Diffusion 1.5 with ControlNet, Stable Diffusion 1.5 with ControlNet and VLM-guidance, the Mixture-of-Experts (MoE) model, and ground truth.}
  \label{fig:qualitative}
\end{figure*}

Figure~\ref{fig:qualitative} presents a qualitative comparison of street-view images synthesized from satellite imagery across three disaster severity levels. Each row corresponds to a damage level, while columns show the results generated by different methods, including Pix2Pix, Stable Diffusion 1.5 with ControlNet, Stable Diffusion 1.5 with ControlNet and VLM-guidance, the Mixture-of-Experts (MoE) model, and the ground truth for reference.

\begin{figure*}[t]
  \centering
  \includegraphics[width=\textwidth]{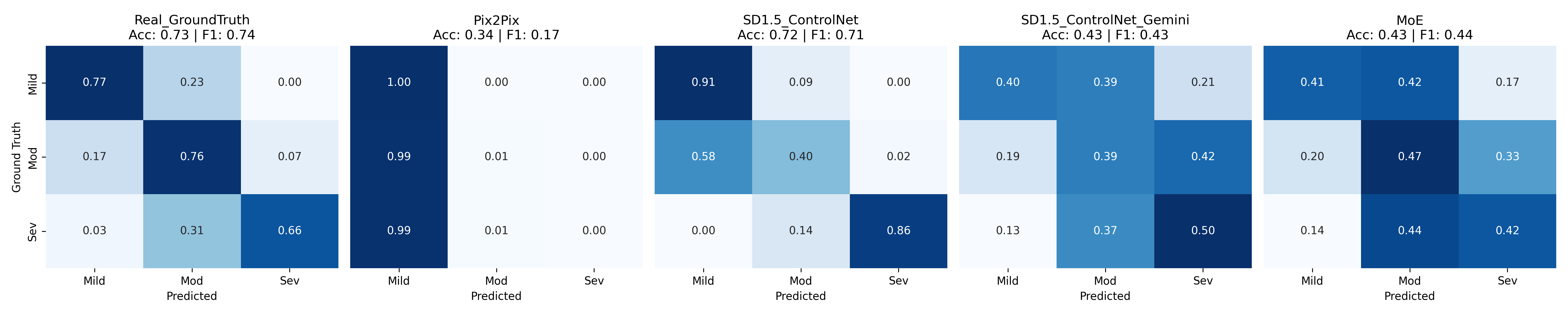}
  \caption{Confusion matrices of disaster severity classification. Pix2Pix collapses to the `Mild' class, while ControlNet shows strong separability. VLM-Guided and MoE models exhibit increased confusion between moderate and severe levels, reflecting the complexity of generated damage patterns.}
  \label{fig:confusion}
\end{figure*}

\begin{table}[t]
\caption{Quantitative Comparison of Perceptual Image Quality Metrics}
\label{tab:iqa}
\centering
\begin{tabular}{lcccc}
\hline
Method & SSIM$\uparrow$ & PSNR$\uparrow$ & LPIPS$\downarrow$ & FID$\downarrow$ \\
\hline
Pix2Pix & \textbf{0.586} & \textbf{15.31} & \textbf{0.549} & 150.83 \\
SD1.5 + ControlNet & 0.314 & 9.81 & 0.602 & \textbf{74.33} \\
SD1.5 + ControlNet + VLM & 0.291 & 9.73 & 0.604 & 82.19 \\
Disaster-MoE & 0.222 & 8.45 & 0.688 & 134.52 \\
\hline
\end{tabular}
\end{table}

Table~\ref{tab:iqa} quantifies the critical trade-off between structural fidelity and perceptual realism. Pix2Pix dominates pixel-level metrics (SSIM: 0.586, PSNR: 15.31), confirming its strict adherence to low-frequency structural layouts. However, its poor perceptual quality is evident in the worst FID (150.83), reflecting the lack of high-frequency textures.

Conversely, diffusion models prioritize perceptual distribution alignment. The SD1.5 + ControlNet achieves the best FID (74.33), indicating superior visual naturalness. However, this realism comes at the cost of geometric precision (SSIM drops to 0.314), consistent with the ``hallucination'' phenomenon observed qualitatively.

Notably, our proposed strategies (VLM-guided and Disaster-MoE) show slightly higher FID and lower SSIM compared to the ControlNet. This performance decline is expected: by explicitly injecting chaotic disaster semantics (e.g., irregular debris), these methods increase pixel-level variance against the ground truth. This limitation of traditional IQA metrics further justifies the necessity of our semantic and perceptual evaluation protocols (Tier 2 \& 3) to assess the actual utility of the generated imagery.

\begin{table}[t]
\caption{Classification Accuracy for Disaster Severity Consistency}
\label{tab:semantic}
\centering
\begin{tabular}{lccccc}
\hline
Method & Acc.$\uparrow$ & F1$\uparrow$ & Mild$\uparrow$ & Mod.$\uparrow$ & Sev.$\uparrow$ \\
\hline
Ground Truth & 0.73 & 0.74 & 0.77 & 0.76 & 0.66 \\
Pix2Pix & 0.34 & 0.17 & \textbf{1.00} & 0.01 & 0.00 \\
SD1.5 + ControlNet & \textbf{0.72} & \textbf{0.71} & 0.91 & 0.40 & \textbf{0.86} \\
SD1.5 + ControlNet + VLM & 0.43 & 0.43 & 0.40 & 0.39 & 0.50 \\
Disaster-MoE & 0.43 & 0.44 & 0.41 & \textbf{0.47} & 0.42 \\
\hline
\end{tabular}
\end{table}

Table~\ref{tab:semantic} reports the semantic consistency evaluated via CAS. The Standard ControlNet achieves state-of-the-art fidelity (F1=0.71), closely matching the Ground Truth upper bound (0.74) and excelling in the \textit{Severe} category (0.86). This confirms that rigid structural constraints effectively preserve discriminative damage features.

In contrast, Pix2Pix suffers from severe mode collapse, trivially achieving 100\% on \textit{Mild} cases but failing on moderate and severe damage classes (F1=0.17). In particular, the VLM-guided and Disaster-MoE methods show a decrease in quantitative consistency (F1 = 0.43 and 0.44, respectively). This aligns with the Realism-Fidelity Trade-off: while these models generate richer textures and debris (as seen in qualitative results), these stochastic details introduce "semantic noise'' that challenges the ResNet classifier, which relies on cleaner structural cues. This result highlights that high perceptual realism does not always correlate with high classification accuracy.

Figure~\ref{fig:confusion} visualizes class-wise discrimination patterns. Pix2Pix exhibits complete mode collapse, classifying almost all inputs as \textit{Mild} regardless of actual damage magnitudes. Conversely, the ControlNet displays a distinct diagonal structure, confirming its superior ability to preserve separable damage features. The VLM-guided and Disaster-MoE models show increased off-diagonal confusion, particularly between \textit{Moderate} and \textit{Severe} classes. This suggests that while these methods enhance perceptual richness (e.g., adding scattered debris), they introduce structural ambiguity that challenges precise classification. This observation reinforces the Realism-Fidelity Trade-off: explicit damage details improve visual naturalness, but may degrade the semantic separability required for automated assessment.

\begin{table}[t]
\caption{Semantic and Perceptual Evaluation by LLM-as-a-Judge}
\label{tab:llm}
\centering
\begin{tabular}{lccc}
\hline
Method & Struct.$\uparrow$ & Damage$\uparrow$ & Realism$\uparrow$ \\
\hline
Pix2Pix & 1.26 & 1.08 & 1.00 \\
SD1.5 + ControlNet & 1.43 & 1.68 & \textbf{2.11} \\
SD1.5 + ControlNet + VLM & \textbf{1.88} & \textbf{2.04} & 2.08 \\
Disaster-MoE & 1.61 & 1.79 & \textbf{2.11} \\
\hline
\end{tabular}
\end{table}

Table~\ref{tab:llm} presents the perceptual evaluation via VLM-as-a-Judge. Pix2Pix scores lowest across all metrics, confirming its inability to synthesize convincing disaster details.

Crucially, this tier reveals the unique value of our proposed methods. While the ControlNet and Disaster-MoE tie for the highest Realism (2.11), validating the superior visual quality of diffusion priors, they differ in semantic correctness. The VLM-guided approach achieves the best Structural (1.88) and Damage Accuracy (2.04).

This result is pivotal: it demonstrates that while standard diffusion models may appear visually realistic, explicit semantic guidance is required to ensure correctness in terms of disaster severity (e.g., collapsed walls and debris). Thus, our VLM strategy effectively bridges the gap between visual hallucination and structural reality, offering the most balanced performance for human-centric assessment.

\section{Conclusion}

This study addresses the scarcity of street-view data in time-sensitive events like post-disaster management by establishing a Structure-Aware Evaluation Framework for cross-view synthesis. Benchmarking four generative paradigms reveals a critical Realism-Fidelity Trade-off: while the Standard ControlNet achieves the best semantic consistency (F1=0.71) and pixel-level fidelity, it often hallucinates structural repairs. Our proposed VLM-guided and Disaster-MoE strategies improve perceptual realism and damage-specific details (as validated by VLM assessments) but introduce stochastic variations that challenge rigid semantic classification. Ultimately, this work highlights the limitations of generating street-view images from satellite observations in post-disaster scenarios using single-model approaches. The results imply that trustworthy disaster generation requires balancing visual plausibility with strict structural alignment, a gap our framework effectively quantifies.

\section*{Acknowledgment}

This work was supported in part by the Texas A\&M University Environment and Sustainability Graduate Fellow Award; the U.S. National Science Foundation (NSF) under Award No. 2318206 (HNDS-I: Cyberinfrastructure for Human Dynamics and Resilience Research); and the Gulf Research Program of the U.S. National Academies of Sciences, Engineering, and Medicine under Grant SCON-10000653. The views expressed are those of the authors and do not necessarily reflect the views of the funding agencies.

\small
\bibliographystyle{IEEEtranN}
\bibliography{references}

\end{document}